%% file: main.tex
\documentclass[10pt,twocolumn,letterpaper]{article}

\usepackage[pagenumbers]{cvpr}      
\usepackage{multirow}
\usepackage{multicol}


\definecolor{cvprblue}{rgb}{0.21,0.49,0.74}
\usepackage[pagebackref,breaklinks,colorlinks,allcolors=cvprblue]{hyperref}

\def\paperID{*****} 
\def\confName{CVPR}
\def\confYear{2025}

\title{GReFEL: Geometry-Aware Reliable Facial Expression Learning under Bias and Imbalanced Data Distribution}

\author{Azmine Toushik Wasi\textsuperscript{1*}, Taki Hasan Rafi\textsuperscript{2*}, Raima Islam\textsuperscript{3}, Karlo Šerbetar\textsuperscript{4}, Dong-Kyu Chae\textsuperscript{2$\dagger$}\\
\textsuperscript{1}Shahjalal University of Science and Technology, Bangladesh
\textsuperscript{2}Hanyang University, South Korea \\ 
\textsuperscript{3}Harvard University, USA 
\textsuperscript{4}University of Cambridge, United Kingdom\\ 
{\small \textsuperscript{*}Co-first authors.
\textsuperscript{\textdagger}Correspondence to:  \tt dongkyu@hanyang.ac.kr}}

\begin{document}
\maketitle
\input{sec/0_abstract}    
\input{sec/1_main_paper}
{
    \small
    \bibliographystyle{ieeenat_fullname}
    \bibliography{main}
}

\input{sec/X_suppl}

\end{document}

%% file: sec/0_abstract.tex
\begin{abstract}
Reliable facial expression learning (FEL) involves the effective learning of distinctive facial expression characteristics for more reliable, unbiased and accurate predictions in real-life settings. However, current systems struggle with FEL tasks because of the variance in people's facial expressions due to their unique facial structures, movements, tones, and demographics. Biased and imbalanced datasets compound this challenge, leading to wrong and biased prediction labels. To tackle these, we introduce \textit{GReFEL}, leveraging Vision Transformers and a facial geometry-aware anchor-based reliability balancing module to combat imbalanced data distributions, bias, and uncertainty in facial expression learning.  Integrating local and global data with anchors that learn different facial data points and structural features, our approach adjusts biased and mislabeled emotions caused by intra-class disparity, inter-class similarity, and scale sensitivity, resulting in comprehensive, accurate, and reliable facial expression predictions. Our model outperforms current state-of-the-art methodologies, as demonstrated by extensive experiments on various datasets.
\end{abstract}

%% file: sec/1_main_paper.tex
 \section{Introduction} \label{sec:intro}
 
One of the most universal and significant ways that people communicate their emotions and intentions is through the medium of their facial expressions \cite{wang2020suppressing}. In recent years, facial expression learning (FEL) has garnered growing interest within the area of computer vision due to the fundamental importance of enabling computers to recognize interactions with humans and their emotional affect states. While FEL is a thriving and prominent research domain in human-computer interaction systems, its applications are also prevalent in healthcare, education, virtual reality, smart robotic systems, etc \cite{ruan2021feature,she2021dive}. 


\begin{figure*}[t] 
\centering {\includegraphics[width=\textwidth]{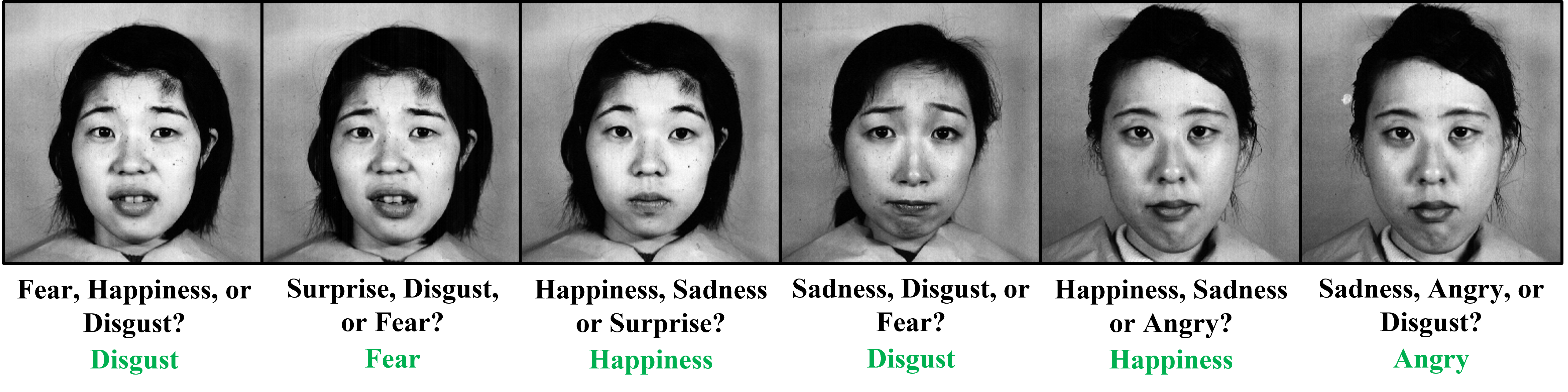}}
\caption{Complexities of Human Emotions (Green-colored labels are true labels).}\label{fig:ExlanationFig}
\end{figure*}

Despite recent strides in facial expression recognition technology, the task remains daunting for several reasons. One major hurdle lies in the diverse and complex nature of human facial expressions (as presented in Figure \ref{fig:ExlanationFig}). People's facial structures, movements, tones, and demographics contribute to a wide variance in expressions, making it challenging for current systems to accurately interpret and classify them.  For instance, telling the difference between a happy smile and a mischievous smirk can be tricky because their lip movements can look quite similar. Also, people express emotions in different ways - some might smile broadly, while others might give a more subtle grin. This variation makes it even harder for computers to accurately pick up on the meaning behind facial expressions. Additionally, consider the challenge of differentiating between a surprised expression and a confused one. Both might involve raised eyebrows and widened eyes, but the context and subtle cues can make a big difference in interpreting the emotion accurately. The complexity of emotions such as anger is often amplified by factors like skin tone and contextual cues, leading to a multitude of potential interpretations that current FEL systems struggle to navigate. \\
These issues, named by intra-class disparity and inter-class similarity, present persistent challenges in facial expression understanding systems \cite{zheng2022poster,Face2Exp,weng2021attentive,mao2023poster}. Within-class variations, such as subtle differences in expression intensity or style, pose difficulties in accurately categorizing similar expressions. For instance, a slight change in eyebrow positioning or mouth curvature can drastically alter the perceived emotion, making classification more ambiguous. Conversely, inter-class similarity adds another layer of complexity, as distinct expressions may share common features or gestures, leading to misclassification. Addressing these nuances is crucial for enhancing the reliability and robustness of FEL frameworks, yet current approaches often fall short in effectively mitigating these challenges. 

Another significant obstacle in FEL stems from biased and imbalanced datasets used for training. These datasets often fail to adequately represent the diversity of facial expressions across different demographics, leading to skewed and inaccurate predictions. For example, datasets may over-represent certain facial expressions commonly exhibited by a particular demographic while under-representing those of others. This imbalance not only undermines the generalizability of FEL models but also perpetuates biases, resulting in erroneous predictions and reinforcing existing societal disparities.


Researchers use several strategies like  unsupervised partitioning, leveraging unlabeled data \cite{ma2023invariant, Face2Exp}, using loss functions \cite{li2022intensity, 9449988afwegg4wag}, ViTs \cite{li2021mvt,mao2023poster,zheng2022poster}, attention-based models \cite{xue2021transfer} and  semi-supervised learning \cite{Li_2022_CVPR}. However, these unsupervised or semi-supervised approaches require extensive additional resources, like large amounts of unlabeled data \cite{98668segsgag25}. Dedicated loss functions for class imbalance may produce harsh results on common labels when prioritizing low-resource classes \cite{horna2ss023deep}. ViTs and attention-based models excel in feature extraction, but may cause poor results in complex emotions with subtle changes \cite{app13045r2225}.
This led us to explore methods tailored to effectively handle diverse facial data on a given dataset. As we know, different facial features can be represented as points in a geometric space \cite{MAJUMDER20141282}, capturing the diverse connections between facial expressions such as lip, nose, eye, and eyebrow movements. These geometric features serve as descriptors for modeling the complexity of facial expressions.

Based on this perspective, we propose a geometry-based reliability balancing system. By placing learnable anchors with center loss to adapt to different facial landmarks and leveraging anchor loss to utilize geometric connections effectively, we aim to capture complex and interconnected emotions effectively. 
We also employ window-based cross-attention ViTs for robust feature learning across facial regions, leveraging their strong capability in feature extraction using both local and global information \cite{mao2023poster}. 
Combining these methods, we introduce a new reliability balancing approach using facial geometry and an attention mechanism. We place anchor points in the embedding space to measure similarity based on facial geometry features and further use multi-head self-attention to identify important features, enhancing the model's reliability and robustness.
This results in improved label distribution and stable confidence scores, mitigating biases and mislabeling caused by various factors. By integrating local and global data using the cross-attention ViT, our approach adjusts for intra-class disparity, inter-class similarity, and scale sensitivity, leading to comprehensive, accurate, and reliable facial expression predictions.

Our contributions are summarized in three folds: 
\begin{itemize}
    \item We propose a novel approach, \textbf{GReFEL}, a novel framework consisting of multi-level attention-based feature extraction with a reliability balancing module for robust FEL with extensive data preprocessing and refinement methods to fight against biased data and poor class distributions.
    \item We introduce geometry-aware adaptive anchors in the embedding space to learn and differentiate between different facial landmarks to increase the reliability and robustness of the model by correcting erroneous labels, stabilizing class distributions for poor predictions, and mitigating the issues of similarity in different classes effectively, addressing intra-class disparity, inter-class similarity, and scale sensitivity.
    \item Empirically, our \textbf{GReFEL} method is rigorously evaluated on diverse in-the-wild FEL databases. Experimental outcomes exhibit that our method consistently surpasses most of the state-of-the-art FEL systems. 
\end{itemize}

\begin{figure*}[t] 
\centering {\includegraphics[width=\textwidth]{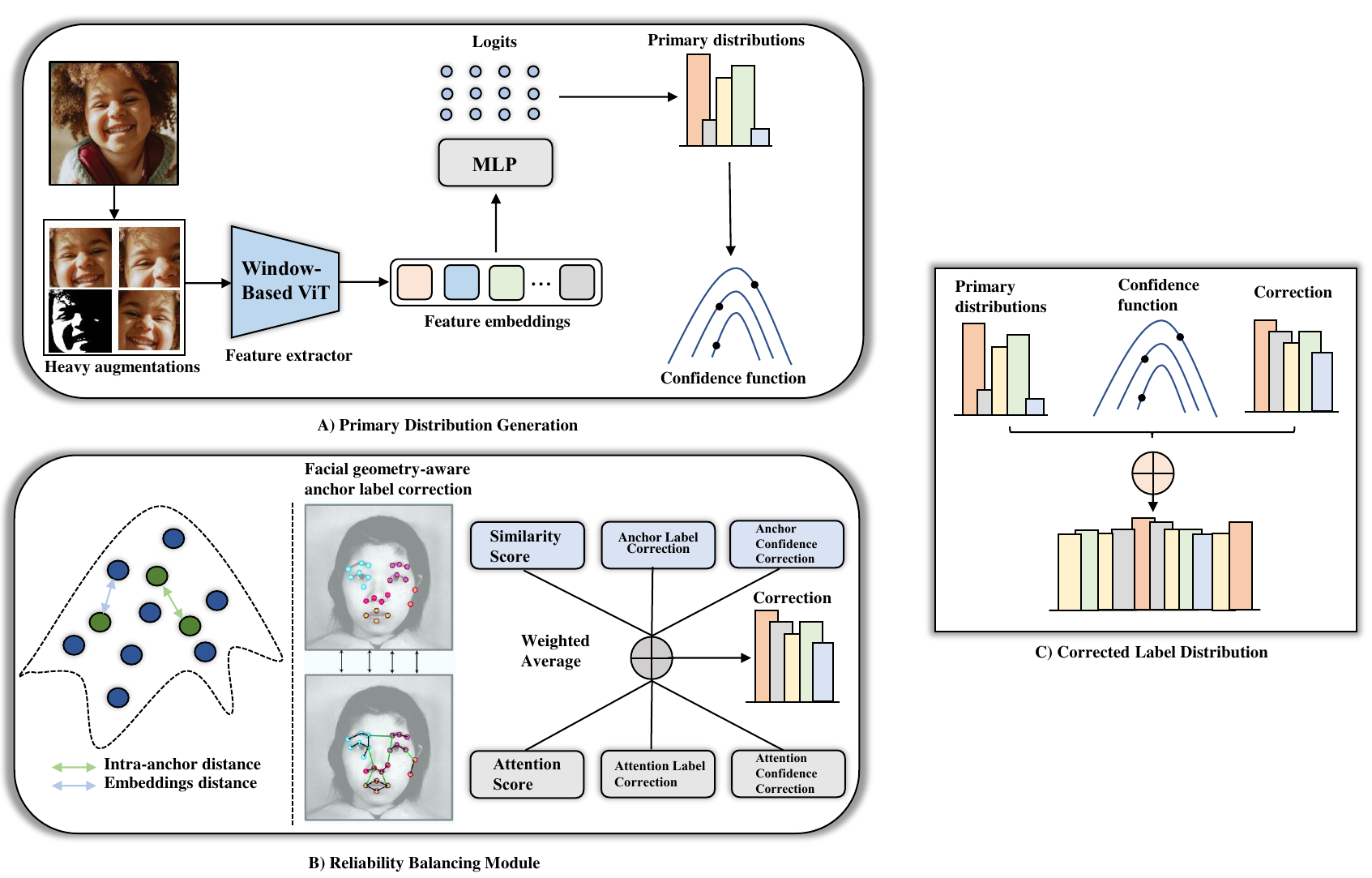}}
\caption{Pipeline of \textbf{GReFEL}. Heavy Augmentation enhances input images, while Data Refinement selects properly distributed class batches per epoch. Window-Based Cross-Attention ViT provides multi-level feature embeddings. MLP predicts primary labels, Confidence is derived from primary label distribution. Reliability balancing utilizes trainable anchors for similarity search and Multi-head self-attention for label correction and confidence calculation. A weighted average of these determines final label correction, resulting in a more reliable model.}\label{fig:2}
\end{figure*}


 \section{Related Works}
 
\noindent \textbf{Facial Expression Learning.} 
Facial expression learning  involves labeling expressions from facial images, comprising facial detection, feature extraction, and expression recognition phases \cite{wang2020suppressing}. Deep learning algorithms, such as self-supervised feature extraction \cite{xue2022coarse}, have optimized FEL systems. Recent advancements include multi-branch networks \cite{weng2021attentive}, uncertainty estimation \cite{she2021dive} and relation-aware local-patch representations \cite{xue2021transfer}. Attention networks based on regions have shown effectiveness for robust FEL\cite{li2018occlusion, wang2020region}.

\noindent \textbf{Vision Transformers in FEL.}
Recent works demonstrate the resilience of Vision Transformers (ViT) against disruption and occlusion \cite{naseer2021intriguing}. Mask Vision Transformer (MVT) addresses FEL challenges by removing complicated backdrops and occlusion, and adapting labels \cite{li2021mvt}. Expression Snippet Transformer (EST) effectively models intra/inter snippet changes for video expression recognition \cite{liu2023expression}. Resilient lightweight multimodal facial expression vision Transformer (MFEViT) handles multimodal FEL data \cite{li2021mfevit}. Neural Resizer balances noise and imbalance in Transformers \cite{hwang2022vision}. Transformer-based multimodal fusion architecture leverages emotional knowledge from diverse viewpoints \cite{zhang2022transformer}. POSTER \cite{zheng2022poster} employs a two-stream pyramid cross-fusion transformer network with a transformer-based cross-fusion method and pyramid structure, while POSTER++ \cite{mao2023poster} simplifies architecture and enhances performance through improved cross-fusion, two-stream design, and multi-scale feature extraction, combining multi-scale features of landmarks with images.

\noindent \textbf{Other Perspectives on FEL.}
Researchers address various challenges in FEL through distinct approaches. INV-REG \cite{ma2023invariant} and Meta-Face2Exp \cite{Face2Exp} reduce data bias using unsupervised partitioning and unlabeled data, respectively. ArcFace \cite{9449988afwegg4wag} and IvReg \cite{li2022intensity} boost discriminative power and dynamic recognition via novel loss functions and attention mechanisms. EAC \cite{EAC_model_zhang2022learn} and Ada-CM \cite{Li_2022_CVPR} handle noisy labels and semi-supervised learning through advanced training strategies. LatentOFER \cite{OFER_10377130} and LA-Net \cite{LA-Net_Wu_2023_ICCV} tackle occlusion and landmark use for improving accuracy and mitigating label noise. M3DFEL \cite{DFER_Wang_2023_CVPR} introduces temporal modeling, while DAN \cite{DAN_biomimetics8020199} captures subtle class differences using feature clustering and attention. Each method contributes unique strategies, reflecting the broad spectrum of challenges and innovations in facial expression recognition. 

Our approach, GReFEL, extracts features using a cross-window-based ViT to get both local and global information, then collects facial landmark geometry data utilizing geometry-aware anchor points and attention mechanisms to learn about distinctive facial data for different emotions effectively, avoiding bias, imbalance, and uncertainties and producing accurate facial expression predictions in real-world scenarios.

\section{Approach}
  
In our approach, we propose a robust feature extraction strategy using ViT and a reliability balancing mechanism to address challenges in FEL. We scale input photos and apply augmentation techniques like rotation and color enhancement for better augmentation. Our pipeline mitigates biases and overfitting by randomly selecting images and expressions during training. Cross-attention ViT is employed for feature extraction, addressing scale sensitivity and intra-class discrepancy. Landmark extraction locates facial landmarks, and a pre-trained image backbone model extracts features. Multiple feature extractors detect low to high-level features, integrated using a cross-attention mechanism for feature vector embedding. Then, primary label distributions are generated using MLPs. Confidence is evaluated using Normalized Entropy. We introduce a reliability balancing method to improve model predictions, addressing limitations in predicting similar classes. Learnable anchors and multi-head self-attention mechanism stabilize label distribution, enhancing reliability. Dropout layers provide additional regularization for robustness against noise and inadequate data. The resulting model, integrating extensive feature extraction and reliability balancing, offers precise and credible predictions even in ambiguous contexts.

\noindent
\textbf{Problem Formulation.}
Let ${x}^i$ be the $i$-th instance variable in the input space $\mathcal{X}$ and $y^i \in \mathcal{Y}$ be the label of the $i$-th instance with $\mathcal{Y} = \{y_1, y_2 \dots y_{N_{cls}}\}$ being the label set.
Let $\mathcal{P}^n$ be the set of all probability vectors of size $n$.
Furthermore, let ${l}^i \in \mathcal{P}^{N_{cls}}$ be the discrete label distribution of $i$-th instance.
Additionally, let ${e} = p({x; \theta_p})$ be the embedding output of the Window-Based Cross-Attention ViT (explained in \ref{Window-Based Cross-Attention ViT}) network $p$ with parameters ${\theta_p}$
and let $f({e}; {\theta_f})$ be the logit output of the MLP classification head network $f_{CH}$ with parameters ${\theta_f}$.

 \subsection{Feature Extraction} \label{Window-Based Cross-Attention ViT}
 
We use a complex image encoder by integrating a window-based cross-attention mechanism, 
to capture patterns from input images.
We extract features by the image backbone and facial landmark detectors. We use IR50 \cite{wang2021face} as image backbone and MobileFaceNet \cite{chen2018mobilefacenets} as facial landmark detector, both pre-trained models. For each level, firstly, division of image features $X_{img} \in \mathcal{R}^{N_p \times D}$ is performed, where $N_p$ represents the number of patches and $D$ denotes the feature dimensions.
The number of patches dictates how the image is fragmented into smaller pieces (e.g., 9 patches would result in $9$ small pieces in \(3 \times 3 \) formation). These patches are then transformed into many non-overlapping windows, \( z_{img} \in \mathbb{R}^{M \times D} \), where $z_{img}$ contains \emph{M} tokens. We use $28\times28$ patches for low-level (local) feature extraction, $14\times14$ for mid-level, and $7\times7$ for high-level (global) feature extraction, as described in Section \ref{sec:ImplementationDetails}.

After $z_{img}\in \mathcal{R}^{M \times D}$, down-sampling of the landmark feature $X_{lm} \in \mathcal{R}^{{A_c} \times H \times W}$ takes place, where ${A_c}$ is the number of channels in the attention network, $H$ and $W$ are the height and width of the image. The down-sampled features are converted into the window size, where the smaller representation of the image is taken and it is represented by $z_{lm}\in \mathcal{R}^{c \times h \times w}$ where $c = D, h \times w$ = M. The features are reshaped in accordance with $z_{img}$’s shape.
The cross-attention with \emph{I} heads in a local window can be formulated as follows at this point:
\begin{equation}
\footnotesize
    q = z_{lm}w_{q}, k = z_{img}w_{k}, v = z_{img}w_{v}
\end{equation} 

\begin{equation}
\footnotesize
    o^{(i)} = softmax(q^{(i)}k^{(i)T}/\sqrt{d}+b)v^{(i)}, i = 1,\ldots,I
\end{equation}

\begin{equation}
\footnotesize
      o = [o^{(1)},\ldots,o^{(I)}]w_o
\end{equation}
where $w_{q}$, $w_{k}$, $w_{v}$ and $w_{o}$ are the matrices used for mapping the landmark-to-image features, and $q, k, v$ denote the query matrix for landmark stream, and key, and value matrices for the image stream, respectively from different windows used in the window-based attention mechanism. [·] represents the merge operation where the images patches are combined to identify the correlations between them and lastly, the relative position bias is expressed as $b \in \mathcal{R}^{I \times I}$ which aids in predicting the placement between landmarks and image sectors.

\begin{figure}[t] 
\centering {\includegraphics[width=\linewidth]{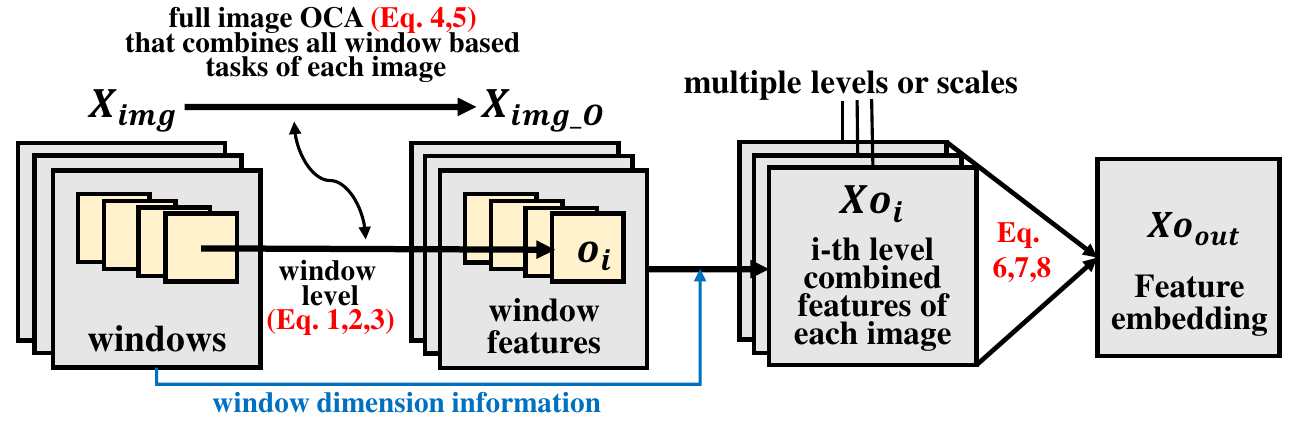}}

\caption{Data flow in the Window-Based Cross-Attention ViT network}\label{fig:WC_ViT}

\end{figure}

We use the equations above to calculate the cross-attention for all the windows, named by \textbf{O}verall \textbf{C}ross \textbf{A}ttention (OCA), as shown in Figure \ref{fig:WC_ViT}. The transformer encoder for the cross-fusion can be calculated as follows:
\begin{equation}
\footnotesize
    X'_{img} = {OCA}_{(img)}+X_{img}
\end{equation}

\begin{equation}
\footnotesize
    X_{img\_O} = MLP(Norm(X'_{img}))+X'_{img}
\end{equation}
where $X'_{img}$ is the combined image feature using OCA, $X_{img\_O}$ the output of the Transformer encoder, and $Norm$(·) represents a normalization operation for the full image of all windows combined. Using window information and dimensions ($z_{img}, M, D, C, H, W, etc.$), we extract and combine window based feature information to $Xo_i$ ($i$-th level window-based combined features of each image) from $X_{img\_O}$ (extracted features of all windows of each image together).

We introduce a vision transformer to integrate the obtained features at multiple scales $Xo_1,...,Xo_i$. Our attention mechanism is able to capture long-range dependencies as it combines information tokens of all scale feature maps:
\begin{equation}
\small
    Xo = [Xo_1,...,Xo_i]
\end{equation}

\begin{equation}
\small
    Xo' = MHSA(Xo) + Xo
\end{equation}

\begin{equation}
\small
    Xo_{out} =MLP(Norm(Xo)) + Xo'
\end{equation}
where [·] denotes concatenation and $MHSA$(·) stands for the multi-head self-attention mechanism. Output of the multi-scale feature combination module $Xo_{out}$, which is equal to feature embedding $e$, is the final output of the encoder network denoted by $p({x; \theta_p})$.

 \subsection{Reliability Balancing} \label{Reliability Balancing with Anchors}
 
Majority of Facial Expression Learning datasets are labeled using only one label for each sample. Inspired by \cite{le2023uncertainty,deng2019arcface}, we provide an alternative approach, in which, we learn and improve label distributions utilizing a label correction approach. We calculate a label distribution primarily that uses the embedding ${e}$ directly into the MLP network. Subsequently, the reliability balancing section employs label correction techniques to stabilize the primary distribution. This results in improved predictive performance through more accurate and reliable labeling. 

\noindent \textbf{Primary Label Distribution. }
From sample ${x}$, using the $p$ network we can generate
the corresponding embedding ${e} = p({x; \theta_p})$ and using the
$f$-network consisting MLP, we can generate the corresponding primary label distribution:
\begin{equation}
\small
{l} = softmax(f({e; \theta_f})).
\end{equation}
We use the information contained in the label distribution with label corrections during training to improve the model performance. 

\noindent \textbf{Confidence Function. }
To evaluate the credibility of predicted probabilities, a confidence function is designed. Let $C_{f}: \mathcal{P}^{N_{cls}} \to [0, 1]$, be the confidence function. 
$C_{f}$ measures the certainty of a prediction made by the classifier using normalized entropy function H($l$). The functions are defined as:

\begin{equation}
\footnotesize 
    C_{f}() = 1 - H({l})
\end{equation}

\begin{equation}
\footnotesize 
    H({l}) = -\frac{\sum_i l^i \log(l^i)}{{N_{cls}}}.
\end{equation}

For a distribution where all probabilities are equal, the normalized entropy is 1, indicating maximum uncertainty, and the confidence value is 0. Conversely, if the probability of one class is 1 and all others are 0, the normalized entropy is 0, indicating no uncertainty, and the confidence value is 1.

 \subsection{Label Correction}
 
The conundrum of label accuracy, distribution stability, and reliability has been a mainstream problem in FEL. The novel approach we propose to resolve this is a combination of two distinct measures of label correction: anchor label correction (geometric) and attentive correction. 

\noindent \textbf{Anchor Label (Geometric) Correction.}
We define anchor ${a}^{ij}$ $(i \in \{1, \dots,$ ${N_{cls}}\},$
$ j \in \{1, 2 \dots K\}$)
to be a point in the embedding space.
Let $\mathcal{A}$ be a set of all anchors.
During training, we use $K$ trainable anchors for each label, with $K$ being a hyperparameter {($k^{th} \in K$)}.
We assign another label distribution ${m}^{ij} \in \mathcal{P}^{N_{cls}}$ to anchor ${a}^{ij}$,
where ${m}^{ij}$ is defined as:
\begin{equation}
m^{ij}_k =
\begin{cases}
    1, \text{ if } k = i \quad (k^{th} \text{anchor}) \\
    0, \text{ otherwise}\\
\end{cases}
.
\end{equation}
Intuitively, here it means anchors ${a}^{1, 1}, {a}^{1, 2} \dots {a}^{1, K}$ are labeled as
belonging to class 1, anchors ${a}^{2, 1}, {a}^{2, 2} \dots {a}^{2, K}$ are labeled as
belonging to class 2 and so on.
To correct the final label and stabilize the distribution, we use geometric information about similarity between the embeddings and  anchors.
The similarity score is $s^{ij}({e})$
is a normalized measure of similarity between an embedding ${e}$ and
an anchor ${a}^{ij} \in \mathcal{A}$.
The distance between  ${e}$ and ${a}$ for each batch and class is:
\begin{equation}
\footnotesize 
    d({e}, {a}) = \sqrt{\sum_{dim_{e}} {|{a} - {e}|^2}}.
\end{equation}
Here, $dim_{e}$ is the dimension of embedding ${e}$. Distances $|{a} - {e}|^2$ are reduced over the last dimension $dim_{e}$ and element--wise square root is taken for stabilizing values. The similarity score $s^{ij}$ is then obtained by normalizing distances:
\begin{equation}
\footnotesize 
    s^{ij}({e}) = \frac{\exp(-\frac{d({e}, {a}^{ij})}{\delta})}{\sum_i^N \sum_j^K \exp(-\frac{d({e}, {a}^{ij})}{\delta})}
\end{equation}
where $\delta$ is a hyperparameter used in the computation of Softmax to control the steepness of the function. The default value used for $\delta$ is 1.0.
From similarity scores we can calculate the anchor label correction term as follows:
\begin{equation}
\footnotesize 
    {t_{g}}({e}) = \sum_i^N \sum_j^K s^{ij}({e}) {m}^{ij}.
\end{equation}

\noindent \textbf{Attentive Correction.}
For multi-head attention \cite{vaswani2017attention}, Let a query with query embeddings $q \in \mathcal{R}^{d_{Q}}$, key embeddings $k \in \mathcal{R}^{d_{K}}$, and value embeddings $v \in \mathcal{R}^{d_{V}}$ is given. With the aid of independently learned projections, they can be modified with $h$, which is the attention head. These parameters are then supplied to attention pooling. Finally, these outputs are altered and integrated using another linear projection. The process is described as follows:
\begin{equation} \label{Eq:16}
\footnotesize 
h_i = f(W_i^{(q)}q, W_i^{(k)}k, W_i^{(v)}v)\in \mathcal{R}^{p_{V}}, W_{out} = W_{o}\begin{bmatrix}h_1 \hdots  h_{n_{heads}} \end{bmatrix}
\end{equation}
where $W_i^{(Q)} \in \mathcal{R}^{d_{model} \times d_Q \textbf{}}, W_i^{(K)} \in \mathcal{R}^{d_{model}\times d_K}$, $W_i^{(V)} \in \mathcal{R}^{d_{model} \times d_V}$, and $W_{o} \in \mathcal{R}^{{n_{heads}}{d_V} \times d_{model}}$ are trainable parameters \cite{vaswani2017attention}, $f$ is the attentive pooling, and each $h_i (i = 1,2,...,n_{heads})$ is an attention head. Also, $d_Q=d_K=d_V=d_{model}/{n_{heads}}$ following \cite{vaswani2017attention}.

As we are using self-attention, all inputs ($q, k, v$ denoting query, key and value parameters respectively) are equal to the embedding ${e}$ \cite{vaswani2017attention}. Self-attention is applied to individual visual embeddings, not across the entire batch. 
${e}$ is passed through the multi-head self-attention layer to obtain the attentive correction term ${t_{a}}$. ${t_{a}}$ is calculated based on the output $W_{out}$ from Eq. (\ref{Eq:16}):
\begin{equation}
\footnotesize 
t_{a} = softmax(W_{out}).
\end{equation}

Multi-head self attention (MHSA) \cite{vaswani2017attention} is designed to focus on the crucial parts relevant to a particular class. Self-attention offers context-aware representations for each sequence element, while multi-head self-attention enhances this by learning various aspects of element relationships, resulting in a more robust understanding \cite{chefer2021generic, vaswani2017attention}. In this work, MHSA can identify important facial areas for each class, thereby improving classification accuracy. 

\noindent \textbf{Final Label correction.}
To combine the correction terms, we use weighted sum, with weighting being
controlled by the confidence of label corrections:

\begin{equation}
\footnotesize 
{t} = \frac{c_g}{c_g + c_a} {t_g} + \frac{c_a}{c_g + c_a} {t_a}
\end{equation}
where $c_g = C_f({t_g})$ and $c_a = C_f({t_a})$. {$t_a$ is the attentive correction term, achieved from $h$ by normalizing.} $C_f()$ stands for the confidence function, calculates confidence of each class predictions.

Finally, to obtain the final label distribution $L_{final}$, we use a weighted sum
of label distribution ${l}$ and label correction ${t}$, as follows:
\begin{equation}
\footnotesize 
    {L_{final}} = \frac{c_l}{c_l + c_t} {l} + \frac{c_t}{c_l + c_t} {t}
\end{equation}
where $c_l = C_f({l})$ and $c_t = C_f({t})$.
The label with the maximum value in the final corrected label distribution $L_{final}$ is provided as a corrected label or a final predicted label.

 \subsection{Loss Function}
 
The loss function used to train the model consists of three terms such as class distribution loss, anchor loss, and center loss. 

\noindent \textbf{Class Distribution Loss ($\mathcal{L}_{cls}$):}
To make sure each example is classified correctly, we use the negative log-likelihood loss between the corrected label distribution ${L_{final}}^i_j$ and label ${y^i}$:
\begin{equation}
\footnotesize 
\mathcal{L}_{cls} = - \sum_i^m \sum_j^N y^i_j \log {L_{final}}^i_j.
\end{equation}

\noindent \textbf{Anchor Loss ($\mathcal{L}_{a}$):}
In order to amplify the discriminatory capacity of the model, we want to make margins between anchors large so that we add an additional loss term:
\begin{equation}
\footnotesize 
\mathcal{L}_{a} = - \sum_i \sum_j \sum_k \sum_l |{a}^{ij} - {a}^{kl}|^2_2.
\end{equation}
We include the negative term in front because we want to maximize this loss. The loss is also normalized for standard uses.

\noindent \textbf{Center Loss ($\mathcal{L}_{c}$):}
To make anchors good representation of their class, we want to make sure anchors and embeddings of the same class stay close in the embedding space.
To ensure that, we add an additional error term:
\begin{equation}
\footnotesize 
\mathcal{L}_c = \min_{k} |{x}^i - {a}^{y^ik}|^2_2.
\end{equation}

\noindent \textbf{Total Loss ($\mathcal{L}_{total}$):}
Our final loss function can be defined as:
\begin{equation}
\footnotesize 
    \mathcal{L}_{total} = \lambda_{cls} \mathcal{L}_{cls}  + \lambda_{a} \mathcal{L}_a + \lambda_{c} \mathcal{L}_c
\end{equation}
where $\lambda_{cls}, \lambda_{a}$, and $\lambda_{c}$ are hyperparameters used to keep the loss functions on the same scale.

\begin{table*}[t]
\small
\centering
\caption{Comparison of Accuracy (\%) ($\uparrow$) with SOTAs. ($^\dagger$in-the-wild datasets  $^*$class-imbalanced)}
\label{tab:Comparison of Accuracy}
\begin{tabular}{l c c c c c c}
\hline
\multirow{2}{*}{Models} & \multicolumn{6}{c}{Datasets} \\ 
\cmidrule(lr){2-7}
 & AffectNet$^{*\dagger}$ & AffWild2$^{*\dagger}$ & RAF-DB$^{*\dagger}$ & JAFFE$^*$ & FER+$^{*\dagger}$ & FERG  \\ \hline
SCN \cite{wang2020suppressing} (CVPR'20) 
& 56.35 & 60.55 & 87.03 & 86.33 & 85.97 & 90.46 \\ 
RAN \cite{wang2020region} (TIP'20)
& 52.97 & 59.81 & 86.90 & 88.67 & 83.63 & 90.22 \\ \hline
DMUE \cite{she2021dive} (CVPR'21)
& 61.21 & 63.64 & 83.19 & - & - & - \\ 
Tr.FER \cite{xue2021transfer} (ICCV'21)
& 66.23 & 68.92 & 90.91 & - & - & - \\ 
RUL \cite{zhang2021relative} (NIPS'21)
& 60.65 & 62.37 & 88.98 & 92.33 & - & 92.35 \\ 
Eff.Face \cite{zhao2021robust} (AAAI'21)
&  59.89 & 62.21 & 88.36 & 92.33 & - & 92.16 \\ \hline
F2Exp \cite{Face2Exp} (CVPR'22)
&  64.23 & 66.34 & 88.54 & - & - & - \\ 
POSTER \cite{zheng2022poster} (IC-W'22)
& 63.34 & 67.74 & 92.05 & 94.57 & 91.62 & 95.82 \\ 
EAC \cite{EAC_model_zhang2022learn} (ECCV'22)
&  61.11 & 63.54 & 88.02  & - & 87.03 & - \\ \hline
L.OFER \cite{OFER_10377130} (ICCV'23)
&  63.90 & 66.02 & 89.60  & - & - & - \\ 
LA-Net \cite{LA-Net_Wu_2023_ICCV} (ICCV'23)
&   64.54 & 66.76 & 91.56  & - & 91.78 & - \\ 
DAN \cite{DAN_biomimetics8020199} (Bioinf.'23)
&   62.09 & 65.82 & 89.70  & - & - & - \\
POSTER$^{++}$ \cite{mao2023poster} ('23)
& 63.76 & 69.18 & 92.21 & 96.67 & 92.28 & 96.36 \\ \hline
\textbf{GReFEL} (Ours) & \textcolor{red}{\textbf{68.02}} & \textcolor{red}{\textbf{72.48}} & \textcolor{red}{\textbf{92.47}} & \textcolor{red}{\textbf{96.67}} & \textcolor{red}{\textbf{93.09}} & \textcolor{red}{\textbf{98.18}} \\ \hline
\end{tabular}
\end{table*}

\section{Experiments}

\subsection{Experimental Setup} \label{sec:ImplementationDetails}

\noindent 
\textbf{Datasets.} \quad
We use \textbf{AffectNet} \cite{Mollahosseini_2019} (420,299 samples; 8 classes), \textbf{Aff-Wild2} \cite{kollias2019affwild2} (1,413,000 samples), \textbf{RAF-DB} \cite{li2017reliable, li2019reliable} (68,718 samples),\textbf{FERG-DB} \cite{aneja2016modeling} (55,769 samples), \textbf{JAFFE} \cite{lyons2020coding} (213 samples), and \textbf{FER+} \cite{BarsoumICMI2016} (35,801 samples) datasets, having 6-8 classes. Among them, AffectNet, Aff-Wild2, FER+, and RAF-DB datasets exhibit class imbalances and are collected in real-world settings.

\noindent 
\textbf{Data Distribution Adjustments.} \quad
We use sample augmentation to expand the training set in class-imbalanced cases, aiding feature identification. Common FEL pre-processing steps include resizing, scaling, rotating, flipping, cropping, color augmentation, and normalization. Uneven class distributions can cause bias and over-fitting. To counter this, equally distributing information from all classes improves model accuracy. Refining datasets ensures balanced training data, mitigating biases. During training, \(N_{pg}\) images are randomly selected from each video or face group. From these, \(B\) images per expression are chosen for training, creating a batch of ($B \times N_{cls}$ (number of classes)) images per epoch, reducing biases and overfitting.

\noindent 
\textbf{Baselines.} \quad
We utilized the following baselines in our experiments: SCN \cite{wang2020suppressing}, RAN \cite{wang2020region}, TransFER (T.FER) \cite{xue2021transfer}, DMUE \cite{she2021dive}, RUL \cite{zhang2021relative}, EfficientFace \cite{zhao2021robust}, Face2Exp (F2Exp) \cite{Face2Exp}, POSTER \cite{zheng2022poster}, EAC \cite{EAC_model_zhang2022learn}, Latent-OFER (L. OFER) \cite{OFER_10377130}, LA-Net \cite{LA-Net_Wu_2023_ICCV}, DAN \cite{DAN_biomimetics8020199}, and POSTER$^{++}$ \cite{mao2023poster}. 

\noindent 
\textbf{Implementation Details.} \quad
For each dataset, we exclusively use cropped and aligned images. These images are resized to 256$\times$256 and then randomly cropped to 224$\times$224 to address overfitting and data imbalance. Heavy augmentation methods are applied during pre-processing as described in Section \ref{sec:ImplementationDetails}. For data refinement, we consider 512 images per video or face group $(N_{pg})$, combining them to create an unbiased set. During training, we select 500 images per class category $(B)$ from this set. {IR50 backbone is trained on Ms-Celeb-1M \cite{guo2016msceleb1m} dataset, MobileFaceNet backbone is trained on Web260M \cite{Zhu_2021_CVPR} dataset, provided via \textit{face.evoLVe} library.} 
Image embeddings are obtained using the Cross Attention ViT network.
In feature extraction, we use $28\times28$ patches for low-level (local) feature extraction, $14\times14$ for mid-level and $7\times7$ for high-level (global) feature extraction.
In Eq. \ref{Eq:16}, $d_Q=d_K=d_V=d_{model}/{n_{heads}}=64$.
Three loss functions are combined for training: Anchor loss maintains distance between anchors, center loss minimizes distance between embeddings and anchors, and class distribution loss ensures correct classification. Our training lasts for 1000 epochs. We employ the ADAM optimizer with an initial learning rate of 0.0003, utilizing exponential decay with $\gamma$ of 0.995 to optimize the model. Primary prediction is done using an MLP with 2 hidden layers of size 64, each followed by ReLU activation, dropout, and batch normalization, except for the last layer. Dropout layers have a drop probability of 0.5 for regularization. For other models, we use default settings as mentioned in their respective papers. 

\subsection{Comparison with State-of-the-Art Methods}
The table \label{tab:Comparison of Accuracy} shows the comparison of the accuracy of multiple State-of-the-Art facial expression learning methods. Upon investigation of the results, it is apparent that GReFEL outperforms all other models across all datasets, attaining the highest accuracy scores for each dataset. Specifically, GReFEL earns an accuracy score of 68.02\% in AffecteNet, 72.48\% on the AffWild2 and 92.47\% on RAF-DB dataset (large in-the-wild dataset), which is significantly higher than POSTER++ and the third best TransFER\cite{she2021dive} (CVPR'21).
Among the compared methods, we think POSTER$^{++}$ (AffectNet 63.76\%, AffWild2 69.18\%) is the most suitable baseline of our work. Compared to this baseline, our ReFEL achieves 68.02\% on AffectNet 72.4\% on AffWild2 (3-5\% better accuracy than POSTER$^{++}$ on these in-the-wild benchmarks).
GReFEL also outperforms every other model in the study, with accuracy scores on the FER+, FERG-DB and JAFFE datasets of 93.09\%, 98.18\% and 96.67\%, respectively, outperforming every other model tested. Our novel reliability balancing section reduces all kinds of biases, resulting in exceptional performance in all circumstances.

\begin{figure*}[t] 
\centering {\includegraphics[width=\textwidth]{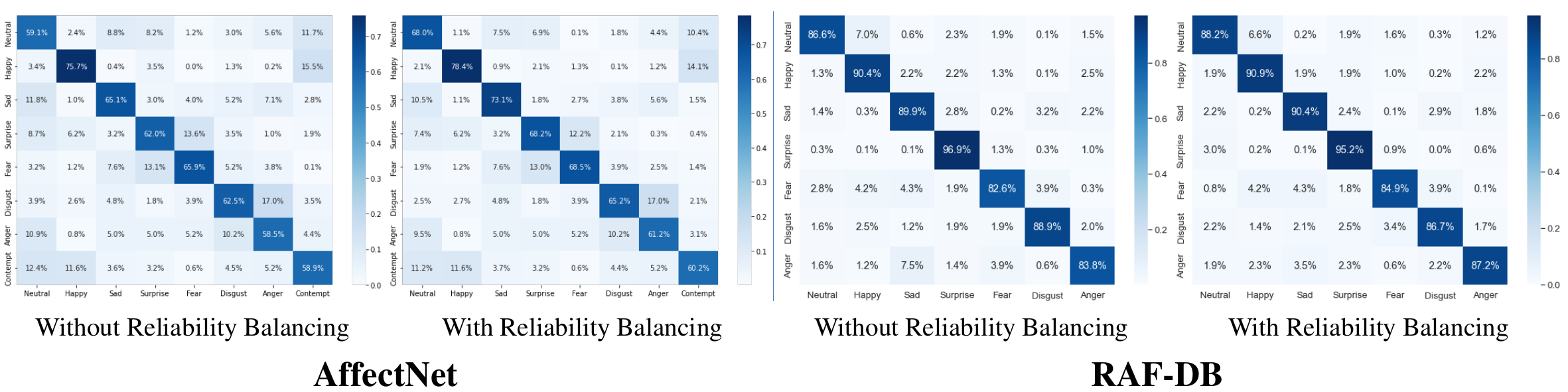}}
\caption{Confusion Matrix.}\label{fig:conf-mat}
\end{figure*}

\begin{figure*}[pht] 
\centering {\includegraphics[width=\textwidth]{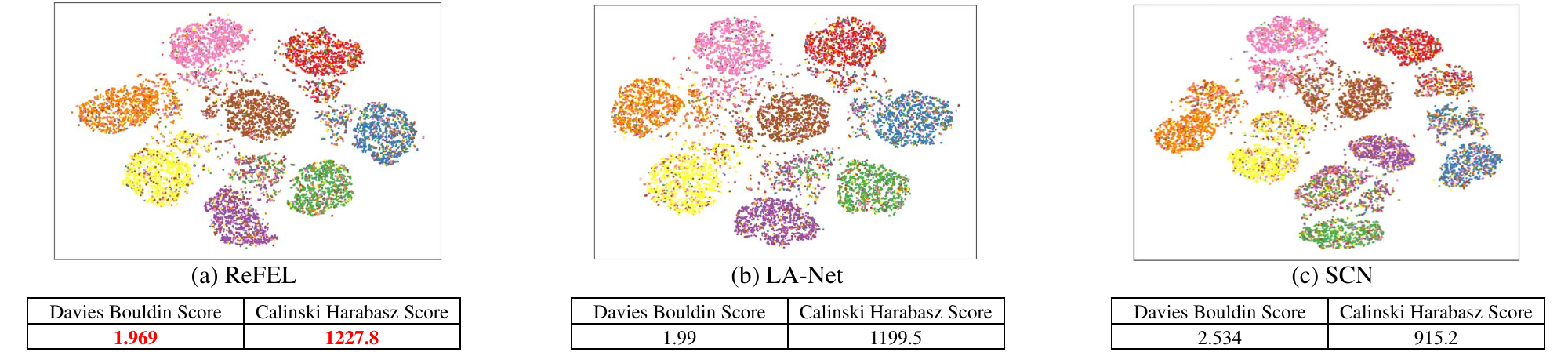}}
\caption{\textbf{t-SNE visualization} of Embeddings with \textbf{Davies Bouldin Score ($\downarrow$) and Calinski Harabasz Score ($\uparrow$)} of our model \textbf{GReFEL} comparing with LA-Net and SCN using \textit{Aff-Wild2} dataset containing 8 classes.}\label{fig:tSNE}
\end{figure*}

\noindent \textbf{Confusion Matrix.} Figure \ref{fig:conf-mat} shows confusion matrices from the AffectNet and RAF-DB datasets, with and without reliability balancing, and reveals several key insights. In AffectNet, reliability balancing notably enhances true positive rates for most emotions, except for neutral and contempt expressions. Without balancing, the classifier struggles with neutral, contempt, and anger distinctions. RAF-DB's performance sees minor improvements with balancing, showcasing a better overall classification compared to AffectNet. Despite this, neutral, contempt, and anger remain challenging to classify accurately. Both datasets show higher true positive rates for surprise and happy expressions, with intriguing confusions between certain emotion pairs like fear and surprise. It indicates that reliability balancing functions effectively, reducing disparities between classes.

\noindent \textbf{Feature Extraction and Clustering.}
In Fig. \ref{fig:tSNE}, the t-SNE plot visually illustrates class differences in the embedding space, with each color representing a distinct class using AffWild2 dataset. The Davies-Bouldin score ($\downarrow$) evaluates cluster resemblance, while the Calinski-Harabasz score ($\uparrow$) measures cluster variance. 
Observations reveal uniformly spaced groups with reliable classifications and noisy areas indicating inter-class similarity and disparity issues. GReFEL outperforms LA-Net and SCN in both Davies-Bouldin (1.969 vs. 1.990 and 2.534) and Calinski-Harabasz scores (1227.8 vs. 1199.5 and 915.2). GReFEL exhibits well-dispersed and discriminating embeddings compared to other models, as evident from the plots and scores.

\subsection{Ablation Study}
Here we explore the impact of different reliability balancing and loss function setups. 
More ablation studies are available in the supplementary materials.

\noindent \textbf{Study of Different Model Setups for Reliability Balancing.} 
The table \ref{tab:modelsetup} summarizes model setups, their accuracy and the F1 score for the AffWild2 dataset. Integration of the Reliability Balancing (RB) module indicates that the F1 scores significantly increase after using reliability balancing methods. We also observe that the initial ViT-based feature extraction requires 43.6M parameters to achieve an accuracy of 68.15\%. However, by incorporating a few additional parameters for reliability balancing, we can significantly enhance the performance, achieving an accuracy of 72.48\% in the model. Also, the increment in computational complexity is minimal.

\begin{table}[h]
\centering
\caption{Reliability Balancing Setups.}  \label{tab:modelsetup}
\begin{tabular}{lcccc}
\toprule
Model Setup  & Accuracy & F1 Score & Params. & FLOPs\\
\midrule
Without RB & 68.15\%  & 0.632 & 43.6M & 8.32G \\
Anchors &  70.36\% & 0.678 & 43.74M  & 8.36G \\
MHSA & 69.05\% & 0.657  & 43.8M & 8.43G\\
Both & 72.48\% & 0.731 & 43.84M & 8.51G\\
\bottomrule
\end{tabular}
\end{table}

\noindent \textbf{Study of Different Loss Setups.} 
Table \ref{tab:loss-setup} summarizes different loss setups and their associated accuracy and F1 score using AffWild2 dataset. Combining classification, anchor, and center losses achieves the highest accuracy of 72.48\%, indicating enhanced model performance through multi-loss integration. More ablations results can be found in the supplementary material.

\begin{table}[t]
\centering
\caption{Loss Setups.} \label{tab:loss-setup}
\begin{tabular}{lcc}
\toprule
Loss  & Accuracy & F1 Score   \\
\midrule
$\mathcal{L}_{cls}$ &  68.15\%  & 0.682 \\
$\mathcal{L}_a $ & 67.96\% & 0.680 \\
$\mathcal{L}_c$ & Not Converge  & Not Converge \\
$\mathcal{L}_{cls}+\mathcal{L}_a$ & 68.87\% & 0.692 \\
$\mathcal{L}_{cls}+\mathcal{L}_a+\mathcal{L}_c$ & 72.48\%  & 0.731 \\
\bottomrule
\end{tabular}
\end{table}

\section{Conclusion}

Our paper introduces \textbf{GReFEL}, a novel FEL approach addressing biased and unbalanced data. GReFEL combines attentive feature extraction with reliability balancing using heavy augmentation and data refinement alongside a Vision Transformer (ViT). Our method effectively handles inter-class similarity, intra-class disparity, and label ambiguity. By incorporating trainable anchor points in embedding space to learn and differentiate between different facial expression landmarks, we stabilize distributions and enhance performance. Experimental analysis across datasets demonstrates GReFEL's superiority over state-of-the-art models, highlighting its potential to advance facial expression learning.

\section*{Acknowledgements} This work was partly supported by (1) the National Research Foundation of Korea(NRF) grant funded by the Korea government(MSIT) (RS-2024-00345398) and (2) the Institute of Information \& communications Technology Planning \& Evaluation(IITP) grant funded by the Korea government(MSIT) (RS-2020-II201373, Artificial Intelligence Graduate School Program (Hanyang University)).

%% file: sec/X_suppl.tex
\clearpage
\maketitlesupplementary

\section{Ablation Studies}

\subsection{Study of Different Values of $\lambda$}
The $\lambda$ values were chosen by our grid search on Aff-Wild2 dataset. Table \ref{tab:lambda} shows the results. Interestingly, setting all $\lambda$ values to 1.0, which is our default setting, achieves the best performance. 

\begin{table}[h] 
\centering
\caption{Experimental results with varying $\lambda$. Only the selected $\lambda$ is modified per experiment, with others set to their optimal values.}
\scalebox{1}{\begin{tabular}{lc|lc|lc}
\toprule
$\lambda_{cls}$  & Accuracy & $\lambda_{a}$  & Accuracy & $\lambda_{c}$  & Accuracy \\
\midrule
0.1  & 35.67\% & 0.1   & 68.18\% & 0.1  & 69.07\%   \\
0.5  & 57.45\% & 0.5   & 69.85\% & 0.5  & 71.02\%    \\
1.0  & \textbf{72.48\%} & 1.0   & \textbf{72.48\%} & 1.0  & \textbf{72.48\%}    \\
\bottomrule
\end{tabular}
}
\label{tab:lambda}
\end{table}

\subsection{Study of Different Loss Functions}
\begin{figure}[h!] 
\centering{\includegraphics[width=\linewidth]{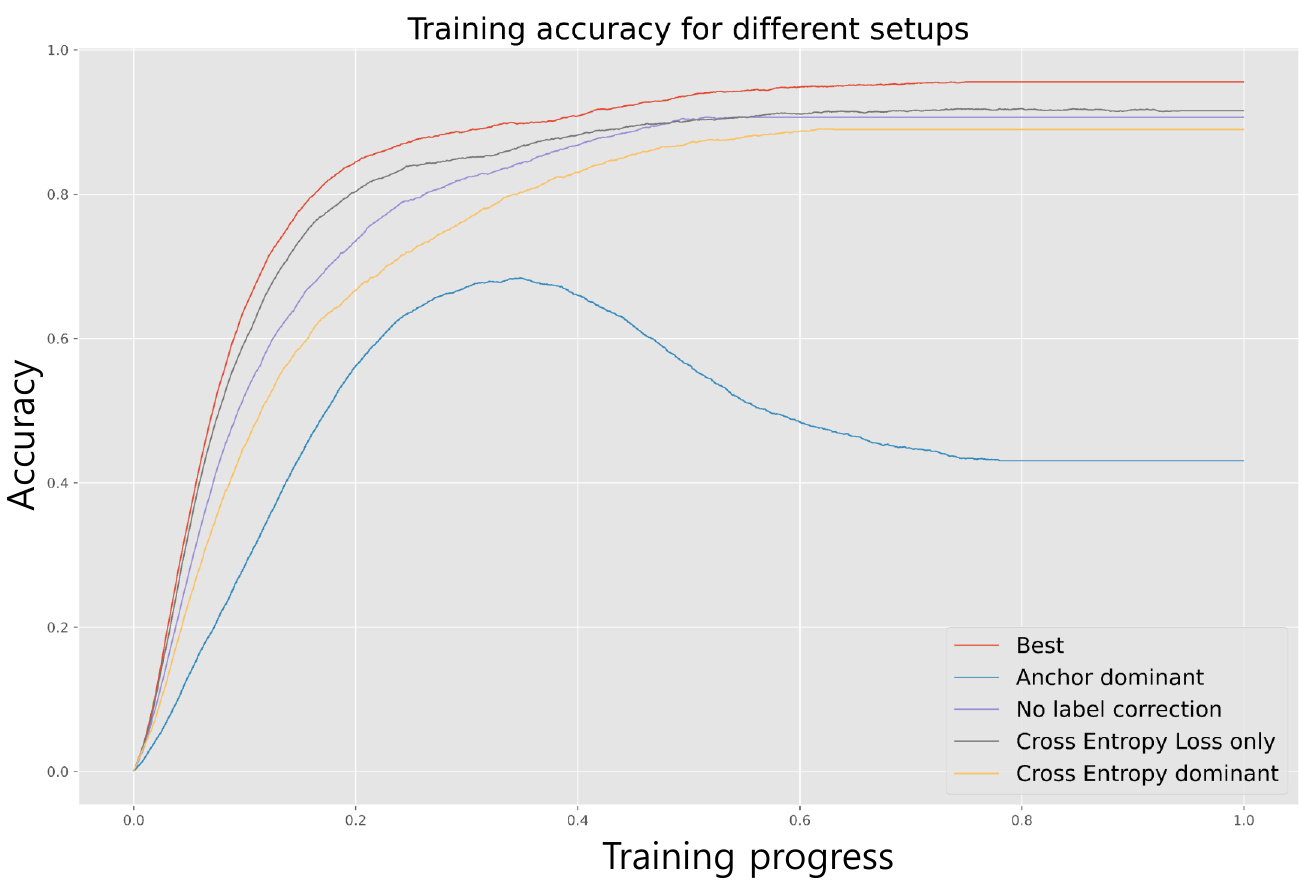}}
\caption{Study of training progress on different setups using Accuracy (\%) score. The red line shows the optimal model with perfect loss combination, blue line shows anchor loss dominant model, indigo colored line shows the model with no label correction with anchors, the gray line shows the model with Cross-Entropy Loss only and the yellow line shows where Cross-Entropy Loss is dominant.}\label{fig:5}
\end{figure}

Fig. \ref{fig:5} demonstrates the effects of different loss function setups in the training stage of our experiment using AffWild2 \cite{kollias2023abaw1} dataset. Anchor loss dominance causes the model to drop its performance after some initial good epochs, conveying that the model starts over-fitting on anchors, ignoring true labels.  Relying more on similarities than the actual prediction performance, this setup fails to fulfill the criteria. The other setups are quite stable and close. The ideal combination used in the study helps the model train faster and better. 

\subsection{Effects of Data Augmentation}
Table \ref{tab:augmentation-noise} shows that without data augmentation, GReFEL sill obtains competitive performance and outperforms POSTER++ in challenging Aff-Wild2 dataset. 

\begin{table*}[h] 
\centering
\caption{Accuracy ($\uparrow$) with and w/o augmentations and noise.}
\scalebox{1}{\begin{tabular}{ccc|ccc}
\toprule
Model & with Augmentation &  w/o Augmentation & Model & 10\% Noise & 0\% Noise\\
\midrule
POSTER++ & 69.18\%    & 66.45\%   & EAC      &  63.54\%    & 64.92\%   \\ 
GReFEL   & \textbf{72.48\%}    & 70.34\%   & GReFEL   &  70.55\%   &  \textbf{72.48\%}  \\
\bottomrule
\end{tabular}
}
\label{tab:augmentation-noise}
\end{table*}

\subsection{Study of Different Number of Anchors \emph{K}}
Table \ref{table:Number of Anchors K vs. Accuracy} demonstrates that optimal recognition accuracy is achieved with 8–10 anchors. Accuracy gradually increases until it reaches this range, beyond which it sharply declines. Few anchors fail to model expression similarities effectively, while excessive anchors introduce redundancy and noise, leading to decreased performance. 

\begin{table*}[h]
\centering
\caption{Number of Anchors \emph{K} vs. Accuracy (\%) ($\uparrow$) means increase in accuracy.} \label{table:Number of Anchors K vs. Accuracy}
\begin{tabular}{c c c c c c c c}
\hline
\textbf{K}  & 0   & 1      & 4  & 6      & 8     & 10    & 20    \\ \hline
\textbf{Accuracy (\%)} & 68.92 & +1.82 ($\uparrow$) &  +2.19 ($\uparrow$) &  +2.26 ($\uparrow$)&\textcolor{red} {\textbf{+2.29}} ($\uparrow$) & \textcolor{red}{\textbf{+2.29}} ($\uparrow$)& +0.51 ($\uparrow$)\\ \hline
\end{tabular}
\end{table*}

\subsection{Study of Noise and Label Smoothing}
\textbf{\emph{K} for Different Noise vs. Accuracy.}
Table \ref{table:Noise vs. Accuracy} illustrates that increasing noise levels decrease model accuracy due to data clarity and complexity issues in AffWild2 \cite{kollias2023abaw1} dataset. However, increasing the value of K improves performance by considering more neighboring points, reducing the impact of noise. Modest yet consistent accuracy improvements are observed with higher K values, but balancing computational complexity is crucial. Over-smoothing from excessively high K values should also be avoided to maintain classification detail.

\begin{table*}[h]
\small
\centering
\caption{\emph{K} for Different Noise vs. Accuracy (\%) ($\uparrow$)} 
\begin{tabular}{c c c c c c c c c c c}
\hline
\emph{K} &  \multicolumn{10}{c}{Noise} \\ \hline
         & 0             & 5             & 10             &  15             & 20             & 25             & 30             & 35             & 40             & 50             \\ \hline
0 & 68.92 & 68.41 & 67.7 & 63.69 & 54.99 & 50.36 & 41.18 & 36.94 & 34.12 & 30.92 \\ \hline
1 & 70.74 & 70.52 & 70.02 & 66.51 & 59.81 & 51.18 & 44.00 & 37.76 & 35.94 & 33.79 \\ \hline
2 & 70.95 & 70.61 & 70.23 & 66.72 & 60.02 & 51.39 & 44.21 & 37.97 & 36.15 & 34.00 \\ \hline
3 & 71.03 & 70.64 & 70.31 & 66.80 & 60.10 & 51.47 & 44.29 & 38.05 & 36.23 & 34.08 \\ \hline
4 & 71.11 & 70.65 & 70.39 & 66.88 & 60.18 & 51.55 & 44.37 & 38.13 & 36.31 & 34.16 \\ \hline
5 & 71.16 & 70.70 & 70.44 & 66.93 & 60.23 & 51.60 & 44.42 & 38.18 & 36.36 & 34.21 \\ \hline
6 & 71.18 & 70.71 & 70.46 & 66.95 & 60.25 & 51.62 & 44.44 & 38.20 & 36.38 & 34.23 \\ \hline
7 & 71.21 & 70.71 & 70.49 & 66.98 & 60.28 & 51.65 & 44.47 & 38.23 & 36.41 & 34.26 \\ \hline
8 & 71.24 & 70.72 & 70.52 & 67.01 & 60.31 & 51.66 & 44.49 & 38.26 & 36.43 & 34.29 \\ \hline
9 & 71.24 & 70.73 & 70.52 & 67.01 & 60.32 & 51.68 & 44.50 & 38.26 & 36.44 & 34.29 \\ \hline
10 & \textcolor{red}{\textbf{71.25}} & 70.73 & 70.53 & 67.02 & 60.33 & 51.69 & 44.51 & 38.27 & 36.45 & 34.3 \\
\hline
\label{table:Noise vs. Accuracy}
\end{tabular}
\end{table*}

\noindent
\textbf{\emph{K} for Different Label Smoothing Terms vs. Accuracy.}
Table \ref{table:Different Label Smoothing Terms vs. Accuracy} illustrates the impact of label smoothing on model accuracy across various K settings in AffWild2 \cite{kollias2023abaw1} dataset. Accuracy generally improves with higher K values, with smoothing terms affecting the degree of improvement. For instance, at K=10, maximum accuracy is 71.89\% with smoothing term = 5, declining to 51.20\% at smoothing term = 40. Smoothing terms between 5 and 20 yield similar accuracy values, making 10 and 11 viable options to balance overconfidence and pattern discovery. A smoothing term of 11 is determined as the optimal choice considering all aspects.

\begin{table*}[h]
\small
\centering
\caption{\emph{K} for Different Label Smoothing Terms vs. Accuracy (\%) ($\uparrow$)} 
\begin{tabular}{c c c c c c c c c c c c c}
\hline
\emph{K} &  \multicolumn{12}{c}{Label smoothing Terms} \\ \hline
  & 0 & 5 & 10 & 11 & 15 & 18 & 20 & 25 & 30 & 35 & 40 & 50 \\ \hline
0 & 68.92 & 69.16 & 69.18 & 69.18 & 69.03 & 68.64 & 67.50 & 64.27 & 61.75 & 59.34 & 55.83 & 50.88 \\ \hline
1 & 70.74 & 71.38 & 71.94 & 71.97 & 71.46 & 70.68 & 70.62 & 67.59 & 63.07 & 60.66 & 56.15 & 51.20 \\ \hline
2 & 70.95 & 71.59 & 72.15 & 72.18 & 71.67 & 70.89 & 70.83 & 67.80 & 63.28 & 60.87 & 56.36 & 51.41 \\ \hline
3 & 71.03 & 71.67 & 72.23 & 72.26 & 71.75 & 70.97 & 70.91 & 67.88 & 63.36 & 60.95 & 56.44 & 51.49 \\ \hline
4 & 71.11 & 71.75 & 72.31 & 72.34 & 71.83 & 71.05 & 70.99 & 67.96 & 63.44 & 61.03 & 56.52 & 51.57 \\ \hline
5 & 71.16 & 71.80 & 72.36 & 72.39 & 71.88 & 71.10 & 71.04 & 67.01 & 63.49 & 61.08 & 56.57 & 51.62 \\ \hline
6 & 71.18 & 71.82 & 72.38 & 72.41 & 71.90 & 71.12 & 71.06 & 67.03 & 63.51 & 61.10 & 56.59 & 51.64 \\ \hline
7 & 71.21 & 71.85 & 72.41 & 72.44 & 71.93 & 71.15 & 71.09 & 67.06 & 63.54 & 61.13 & 56.62 & 51.67 \\ \hline
8 & 71.24 & 71.86 & 72.44 & 72.47 & 71.96 & 71.18 & 71.12 & 67.09 & 63.57 & 61.16 & 56.65 & 51.70 \\ \hline
9 & 71.24 & 71.88 & 72.44 & 72.47 & 71.96 & 71.18 & 71.12 & 67.09 & 63.57 & 61.16 & 56.65 & 51.70 \\ \hline
10 & 71.25 & 71.89 & 72.45 & \textcolor{red}{\textbf{72.48}} & 71.97 & 71.19 & 71.13 & 67.1 & 63.58 & 61.17 & 56.66 & 51.71 \\ \hline
\label{table:Different Label Smoothing Terms vs. Accuracy}
\end{tabular}
\end{table*}

\subsection{Study of Primary Mislabeled Predictions}
\begin{figure}[h] 
\centering {\includegraphics[scale=0.5]{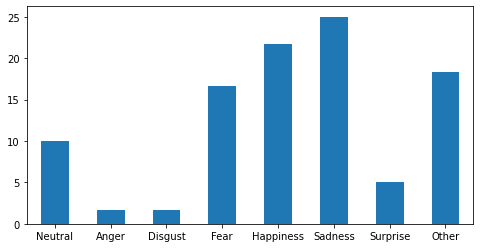}}
\caption{Percentage of incorrect labels among all incorrect labels in the AffWild2 dataset for GReFEL}\label{fig:label_change_GR}
\end{figure}

Figure \ref{fig:label_change_GR} illustrates the proportion of mislabeled images among all mislabeled instances using the AffWild2 dataset. Notably, happiness, sadness, and fear exhibit the highest mis-prediction rates, followed by other and neutral emotions. These trends can be attributed to the intricate nature of certain emotions discussed in introduction section of the main paper. Distinguishing subtle variations between happiness and surprise, or between sadness and neutral states, poses challenges for accurate prediction; and our model effectively solves the issue.

\begin{figure}[h] 
\centering {\includegraphics[scale=0.5]{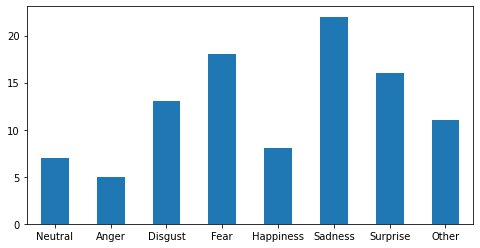}}
\caption{Percentage of incorrect labels among all incorrect labels in the AffWild2 dataset for SCN}\label{fig:label_change_reb}
\end{figure}

We have compared label correction of ours with SCN on the AffWild2 dataset. Figure \ref{fig:label_change_reb} shows the result of SCN. For SCN, the errors are higher for \textit{Surprise}, \textit{Anger} and \textit{Disgust} more than GReFEL, indicating a more robust feature extraction of GReFEL. Additionally, GReFEL performs better with complex and ambiguous emotions such as \textit{Anger}, \textit{Disgust}, and \textit{Fear} when compared to SCN.

\section{Explaining Reliability Balancing}
\begin{figure*}[h] 
\centering {\includegraphics[width=\linewidth]{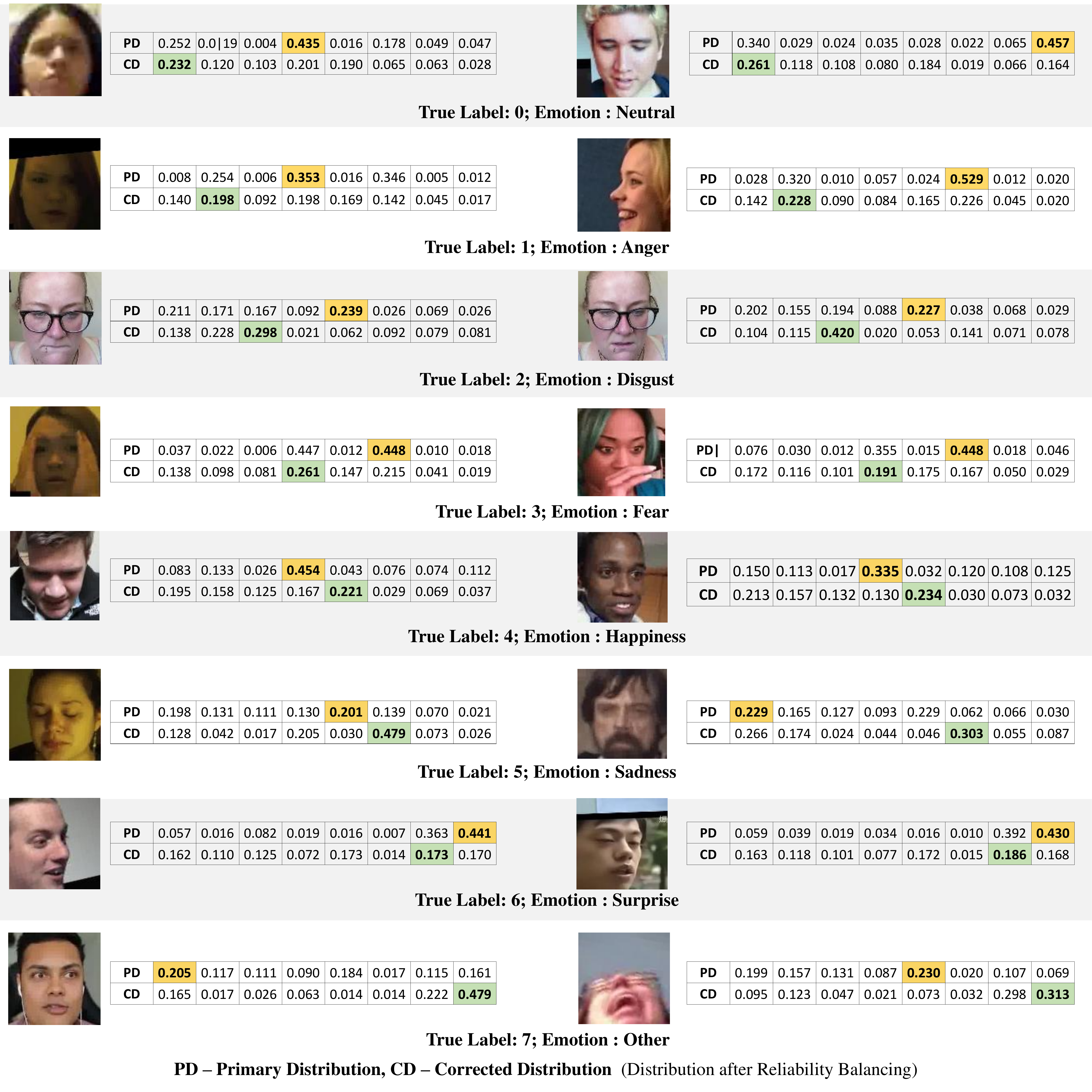}}
\caption{Observation of \textbf{confidence probability distributions in GReFEL} using \textit{Aff-Wild2} dataset. Eight different emotions—\textbf{Neutral, Anger, Fear, Disgust, Happiness, Sadness, Surprise,} and \textbf{Other}—are represented by columns under each image sequentially. \textbf{Primary Distribution (PD)} is the initial prediction, while \textbf{Corrected Distribution (CD)} is the accurate prediction after \textbf{Reliability Balancing}. The correct label after reliability balancing is marked as green, and the inaccurate primary prediction label is marked as yellow.
}\label{fig:3}
\end{figure*}

The reliability balancing module plays a crucial role in enhancing the accuracy and reliability of predictions by stabilizing probability distributions in our framework. This strategy increases probability confidence values for appropriate labels while decreasing confidence in incorrect predictions, as Fig. \ref{fig:3} clearly indicates. For instance, Labels 2, 5, and 7 experience a noticeable rise in their maximum confidence values after applying reliability balancing, ensuring more accurate predictions. Conversely, the method reduces the confidence levels of incorrect predictions, as seen in Labels 0, 1, and 3, where the incorrect maximum values decrease to a range of 0.15-0.25. Notably, even in these cases, the correct labels maintain a probability range of 0.2–0.3, enabling the model to make the right predictions.
After implementing the corrective measures, the maximum and minimum probabilities across the sample increased to 0.5429 and 0.0059, respectively, resulting in a more stable and balanced distribution. A key observation is that the standard deviation of the corrected predictions (0.0881) was found to be lower than that of the primary predictions (0.1316), providing strong evidence for enhanced stability and balance.

Furthermore, the reliability balancing strategy proves invaluable in scenarios where the primary model struggles with label ambiguity, intra-class similarity, or disparity issues within the images. As evident from Fig. \ref{fig:3}, even when the maximum primary probability exceeds 0.4, the associated labels may be erroneous, rendering the model unreliable. Thus, the reliability balancing method supports the model in both extremely uncertain conditions and extremely confident scenarios where the primary model makes poor conclusions.

